\documentclass[letterpaper, 10 pt, conference]{ieeeconf}  

\IEEEoverridecommandlockouts                              
\usepackage{graphics} 
\usepackage{graphicx}
\usepackage{epsfig} 
\usepackage{mathptmx} 
\usepackage{times} 
\usepackage{amsmath} 
\usepackage{amssymb}  
\usepackage{cite}
\usepackage{url}
\usepackage{marvosym}
\usepackage{epstopdf}

\title{\LARGE \bf
DoraPicker: An Autonomous Picking System for General Objects 
}

\author{Hao Zhang, Pinxin Long, Dandan Zhou, Zhongfeng Qian, Zheng Wang, \\ Weiwei Wan, Dinesh Manocha, Chonhyon Park, Tommy Hu, Chao Cao, Yibo Chen, Marco Chow, Jia Pan
\thanks{Hao Zhang and Dandan Zhou are with Dorabot Inc., Shenzhen, China; Zhongfeng Qian, Zheng Wang, Tommy Hu, Chao Cao, Yibo Chen, and Marco Chow are with the University of Hong Kong, Weiwei Wan is with Advanced Industrial Science and Technology, Japan; Dinesh Manocha and Chonhyon Park are with the University of North Carolina at Chapel Hill; Pinxin Long and Jia Pan are with the City University of Hong Kong. {\tt\small info@dorabot.com; jiapan@cityu.edu.hk}
}
}

\begin{document}
\maketitle
\thispagestyle{empty}
\pagestyle{empty}

\begin{abstract}
Robots that autonomously manipulate objects within warehouses have the potential to shorten the package delivery time and improve the efficiency of the e-commerce industry. In this paper, we present a robotic system that is capable of both picking and placing general objects in warehouse scenarios. Given a target object, the robot autonomously detects it from a shelf or a table and estimates its full 6D pose. With this pose information, the robot picks the object using its gripper, and then places it into a container or at a specified location. We describe our pick-and-place system in detail while highlighting our design principles for the warehouse settings, including the perception method that leverages knowledge about its workspace, three grippers designed to handle a large variety of different objects in terms of shape, weight and material, and grasp planning in cluttered scenarios. We also present extensive experiments to evaluate the performance of our picking system and demonstrate that the robot is competent to accomplish various tasks in warehouse settings, such as picking a target item from a tight space, grasping different objects from the shelf, and performing pick-and-place tasks on the table. 
\end{abstract}

\section{Introduction}
\label{sec:intro}
Modern electronic commerce companies such as Amazon are able to package and ship millions of items to end customers from a network of fulfillment centers all over the globe. This would not be possible without leveraging cutting-edge techniques in robotics, including perception, planning, grasping and manipulation~\cite{baker2007exploration}. For instance, Amazon's automated warehouses are successful at removing much of the human workload in searching for items within a warehouse~\cite{wurman2008coordinating}. In China, electronic commerce companies also heavily depend on intelligent supply-chain management services provided by companies like BlueSword Logistics. However, commercially feasible automated picking in unstructured or semi-structured environments still remains a difficult challenge, and it is also important for many other applications such as manufacturing and service robots. Generalized picking in cluttered environments requires successful integration of many components, such as object recognition, pose estimation, grasp planning, compliant manipulation, motion planning, task planning, task execution, and error detection/recovery.

In this paper, we present an autonomous robotic system -- DoraPicker -- that is able to perform picking and placement operations in a simplified warehouse scenario. The prototype of this system participated in the Amazon Picking Challenge (APC) in 2015~\cite{apc2015}. In the APC, teams need to build an autonomous system for performing a task that is currently achieved manually by human workers in the warehouse, i.e., picking specific objects from shelves and placing those objects into specified containers. The shelves were prototypical pods provided by the competition organizers, and the objects were selected from the popular set of items sold on Amazon. These objects are placed on a variety of poses and positions in the shelves. 
In addition, the environment surrounding the robot, e.g., the specification of the shipping box (container), the relative position to the shelf and other static obstacles, are all known in advance. By combining a series of robotics techniques, the DoraPicker is able to accomplish this warehouse task robustly and efficiently. 

\begin{figure}
\includegraphics[width=\linewidth]{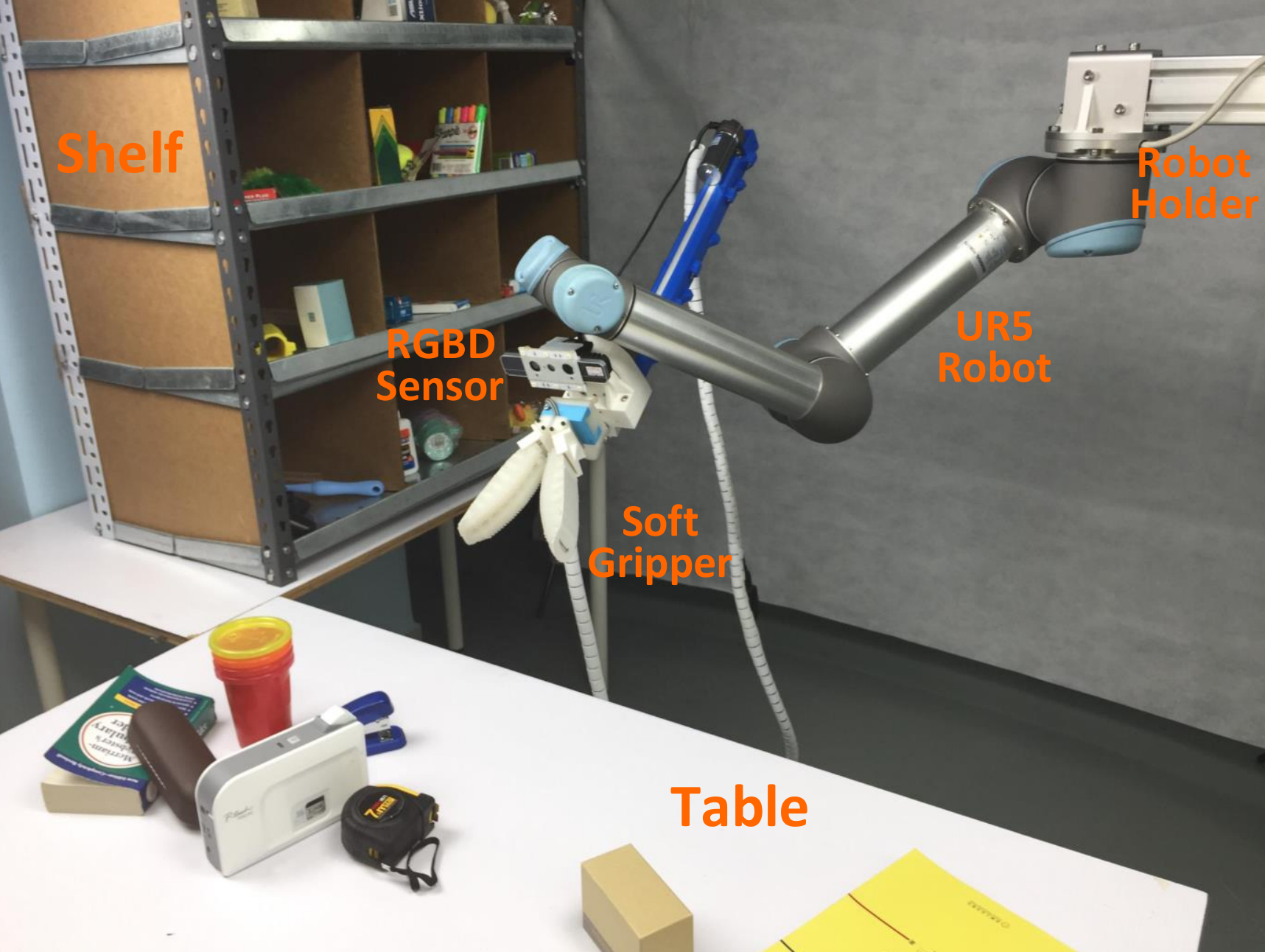}
\caption{The DoraPicker autonomous picking system.} 
\label{fig:dorapicker}
\end{figure}

\begin{figure}
\centering
\includegraphics[width=1.0\linewidth]{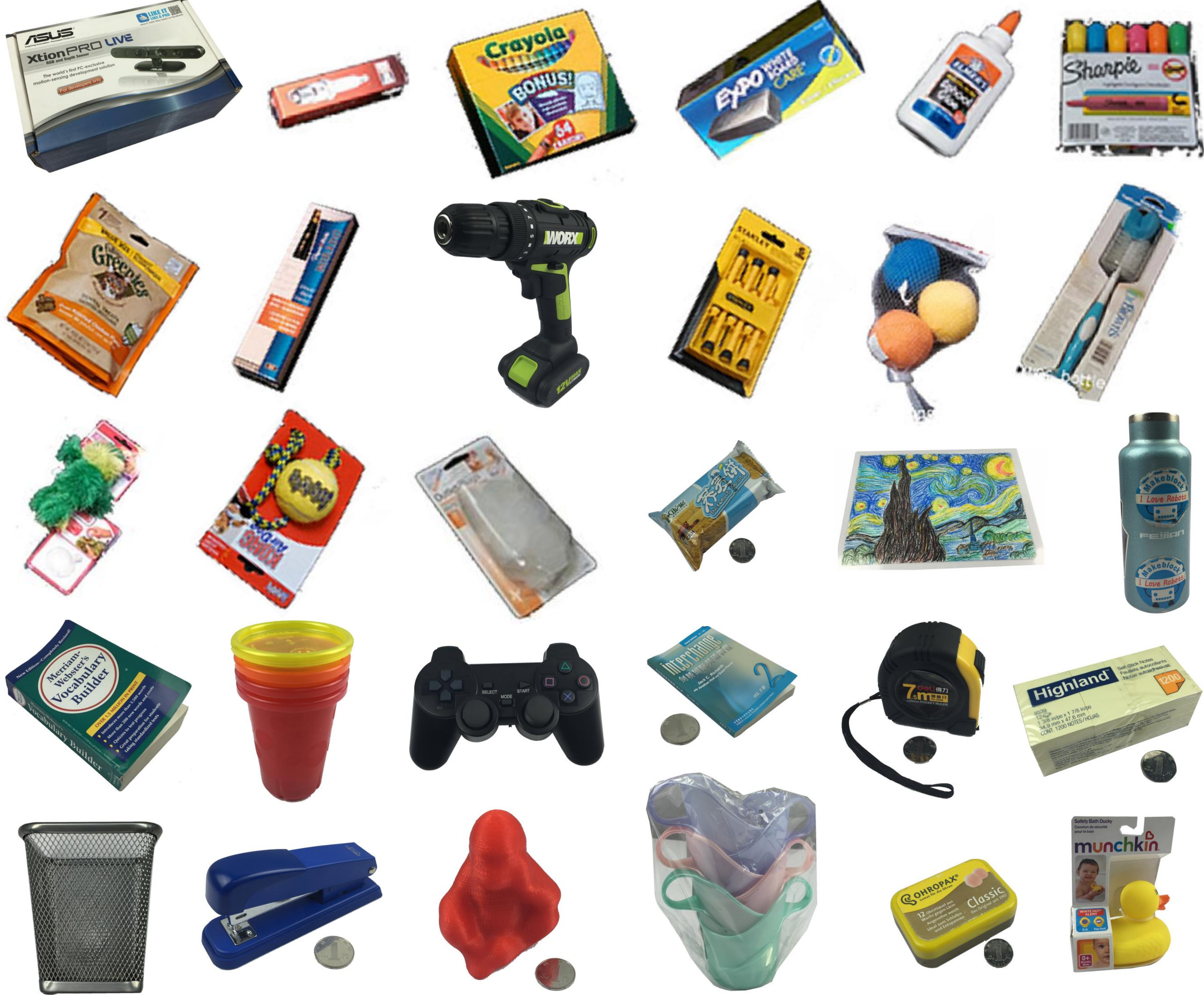}
\caption{The 30 objects used in our experiments. Row-by-row starting from the top-left: Cardboard box, spark plug, Crayola crayons, whiteboard eraser, glue, sharpies, cat treats, pencils, electric drill, screwdrivers, foam balls, bottle cleaner, dog toy - frog, dog toy - ball, outlet protectors, cookie, picture, bottle, dictionary, plastic cups, gamepad, little book, pocket ruler, sticky notes, pencil holder, stapler, 3D printed model, set of cups with package, earplug box, rubber ducky.}
\label{fig:objects}
\end{figure}

Our system takes a list of target object names as the input, then retrieves the target item from the shelf or the table and estimates its 6D pose related to the robot. After that, motion planners calculate a collision-free trajectory for the robot to approach the item. Following this trajectory, the robot moves towards the target object, grasps it using the gripper, and then places the object into a specified area. In order to pick general objects with a large variety in terms of shape, weight and material, we designed three grippers, including a vacuum gripper, an Openhand M2 gripper~\cite{ma2016m2} and a soft gripper. In the experiment, we demonstrate the performance of our picking system on two scenarios, one is similar to the APC shelf workspace, and the other is a table scenario where the robot need to perform picking-and-placement tasks on a tabletop setting. We also evaluate the performance of the three grippers on a set of objects with different color, geometry, material and surface characteristics (as shown in Figure~\ref{fig:objects}). The experiments show that the soft gripper can grasp most of the tested objects, the vacuum gripper is apt for picking up the lightweight objects from the shelf, and the Openhand M2 gripper has robust performance on pick-and-place tasks but it is not suitable for picking objects from the shelf.

The paper is organized as follows: Section~\ref{sec:related} provides a brief survey over the related work. In Section~\ref{sec:overview}, we give an overview of our autonomous robot, including hardware components and the software architecture. Section~\ref{sec:perception} and~\ref{sec:planning} describes perception and planning part of our system respectively. Section~\ref{sec:gripper} presents three grippers designed for picking up objects. Section~\ref{sec:experiment} then contains some preliminary results and Section~\ref{sec:lesson} and Section~\ref{sec:conclusion} concludes the paper.

\section{Related Work}
\label{sec:related}
Recently, warehouse automation has received more and more attention in the research community~\cite{baker2007exploration}. While the Kiva system\footnote{Now it is renamed as Amazon Robotics.} acquired by Amazon in 2012 has been a success in making Amazon's warehouses more productive~\cite{guizzo2008warehouse}, robots that can autonomously grasp items from the shelf is still a challenging and active research topic. 

Previous work on warehouse automation mainly focuses on autonomous transport, e.g., delivering the packages by using Automated Guided Vehicles (AGV)~\cite{garibotto1998industrial}~\cite{barbera2003fork}. Beyond that, in~\cite{cosma2004autonomous}, it describes an autonomous robot for indoor light logistics with limited manipulating ability. This robot is designed for transporting packages to the destination in a pharmaceutical warehouse. Although it is able to pick up the packages from the ground, its manipulation capability is limited and not suitable for the e-commerce application, which requires grasping various objects from shelves and tables. Manipulation in restricted spaces like boxes and shelves leads to difficult high-dimensional motion planning problems. To this end,~\cite{pan2012collision} proposed a sample-based motion planning algorithm that performs local spline refinement to compute smooth, collision-free trajectories and it works well even in environments with narrow passages. 

Another related area to our system is bin-picking, which addresses the task of automatically picking isolating single items from a given bin~\cite{harada2013probabilistic}~\cite{liu2012fast}~\cite{nieuwenhuisen2013mobile}~\cite{domae2014fast}. However, in the bin-picking tasks, the working space of the robot is only the given bin, which is a structured, limited and relatively easy operation area compared to our warehouse settings including a table and a shelf with 12 bins. 

The perception problem in the warehouse settings is an example of the general object detection, segmentation and pose estimation, which is widely researched in computer vision. But this perception problem also has its own characteristics, i.e., it typically uses multi-modal vision (not only the RGB images) and tightly couples with the following grasping movement~\cite{wan2016teaching}. To accurately manipulate the target object, our perception methods need to output the object's full 6D pose. Traditional computer vision methods usually output bounding boxes with highly likely object locations on the input RGB images. These representations are difficult to use in warehouse picking context as the bounding boxes without depth information are not that useful for grasping and manipulating. Recently, there are some work on creating RGB-D datasets~\cite{bird,rennie2015dataset} for improving object detection and pose estimation in warehouse environments. Specially, the multi-class segmentation used in~\cite{multiclass} helps the team achieve the winning entry to the APC 2015. This method gives shape information as output to indicate the location of the target object. Although shape information is sufficient to pick items in some situations by using vacuum grippers, it is inadequate for dexterous grippers to accurately manipulate objects. 


\section{Overview}
\label{sec:overview}
The picking system DoraPicker (as shown in Figure~\ref{fig:dorapicker}), is a stationary manipulator that consists of a 6-DOF UR5 robotic arm, a holder for fastening the robot, several different grippers that can grasp various objects and an Intel Realsense RGBD sensor for acquiring the environmental information.

\subsection{Workspace Setup}
We set up our picking system in a simplified warehouse setting as shown in Figure~\ref{fig:dorapicker}. This simplified scenario contains two representative operating environments in common warehouses, i.e, a shelf and a table. The robot not only grasps an item from the shelf and releases the item into a container, but also performs the pick-and-place task on the table. 

For manipulating objects in all bins of the shelf (or on different locations of the table), it is important to figure out an optimized mounting pose of the robot relative to the shelf and the table. We determine this pose through computing the manipulability of our system based on specification of the UR5 robot. Then we calibrate the robot against the shelf and the table mechanically. Since the relative position between the robot and its workspace is known, we can estimate a quite accurate pose of the shelf and the table based on the robot's joint angle and forward kinematics.

\subsection{The Robot and The RGBD Camera}
The UR5 robot has 6 DOF and each DOF can rotate 720 degrees. It has a 1.5kg payload. We use the ur\_driver package\footnote{\url{http://wiki.ros.org/universal_robot}} from ROS-Industrial as the interface to communicate with the UR5 robot. To broaden the reachability of the UR5 robot, we add a prismatic joint to its end-effector, that is to say, the entire system has 7 DOF. This prismatic joint can maximally extend 35cm, which enables our grippers to approach the deepest part of the shelf. 

In order to obtain relative high-resolution RGB and depth information, we mount an Intel Realsense RGBD camera to the end-effector of UR5. This camera has a full 1080p RGB sensor at 30fps and a 640x480 depth resolution at 60fps. The depth sensor in this camera has 0.2$\sim$1.2 meter range which is appropriate for the object-level perception. We have developed a ROS package as an open-source driver\footnote{\url{http://wiki.ros.org/realsense_camera}} for the camera. This package can generate real-time colored point clouds from raw depth map and RGB image utilizing existing UV map from Intel Realsense's SDK\footnote{\url{https://software.intel.com/en-us/intel-realsense-sdk}} in Windows. 

To calibrate the mechanical relation between camera and the robot, we exploited the VISP~\cite{marchand2005visp} software platform and the industrial\_extrinsic\_cal package\footnote{\url{http://wiki.ros.org/industrial_extrinsic_cal}} from ROS-Industrial for the hand-eye calibration.

\subsection{Software Architecture}
\begin{figure*}
\centering
\includegraphics[width=1.0\linewidth]{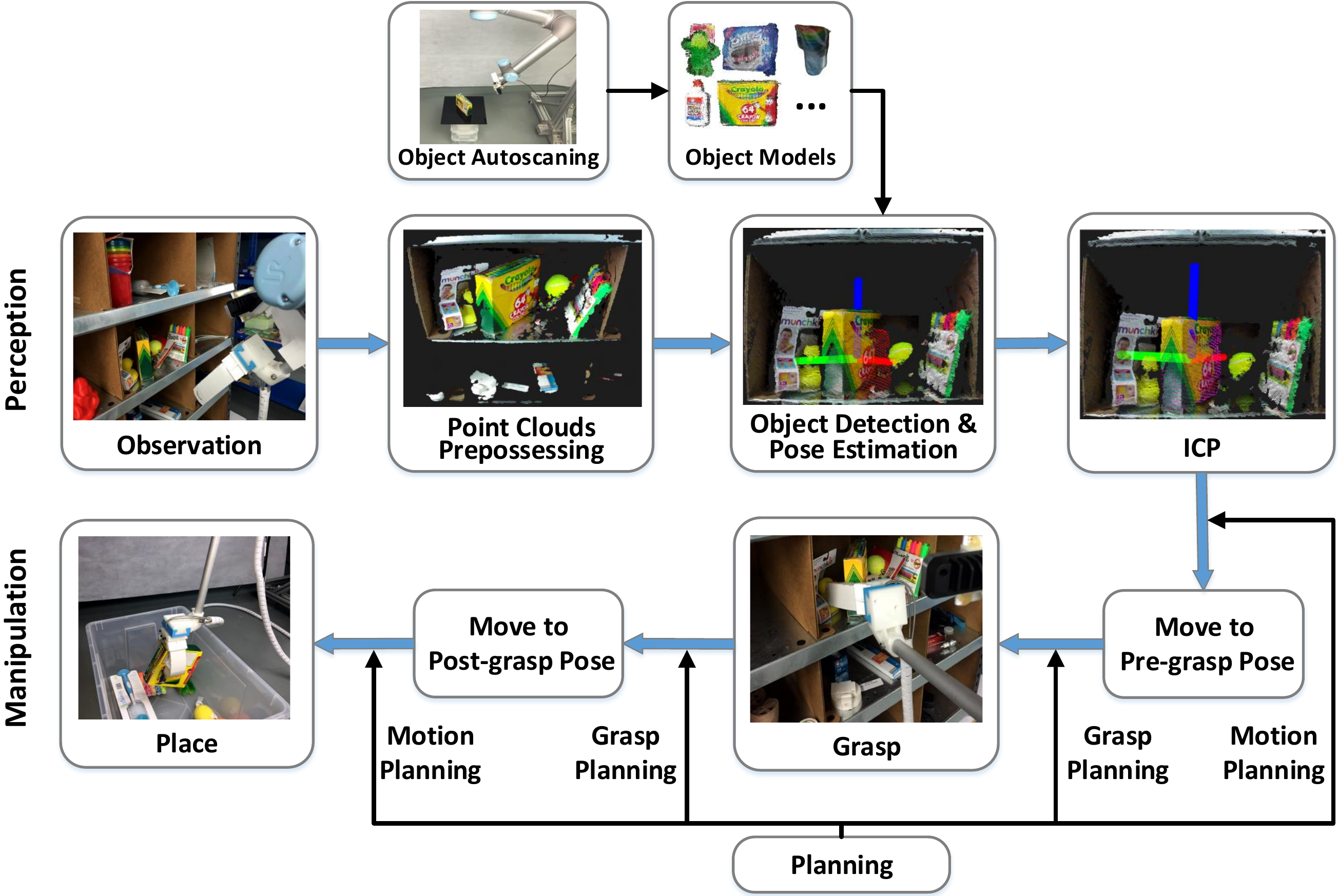}
\caption{The software architecture of DoraPicker.}
\label{fig:overview}
\end{figure*}

Figure~\ref{fig:overview} illustrates the software architecture of the DoraPicker system. The input to our system is a list provided by the user that contains the names of desired objects. Once received the list, the system parses it and moves the robot to an observation pose predetermined for each bin based on the known information of the shelf (the same method is used to define the observation pose for the table). At the observation pose, the RGBD camera will acquire the depth-registered point clouds of the target object. If the robot cannot detect the target object based on the current point clouds, it will slightly move the end-effector to the sides of the observation pose and acquire the input again. If it fails to find the target object at all three poses (the observation pose and its two sides), the robot will move away to pick the next object in the list. Before we detect the target object from the sensory input, two preprocessing methods are used to filter them to noise-free, low-density point clouds. With the preprocessed point clouds, we perform the template-based object detection and pose estimation to locate the target object in the workspace. Then the iterative closest point (ICP)~\cite{icp} method is used to register the resulting template to the target object's point cloud to get more accurate 6-DOF pose information. 

Given the pose of the target item, the robot arm moves to a pre-grasp pose along the collision-free trajectory generated by motion planners. Considering the relative pose between the target item and end-effector at the pre-grasp pose, the grasp planner will choose a valid way from a predefined grasping database for the gripper to approach the target item. The robot extends its gripper and executes the selected grasp plan. To place the picked item to a specified container, we first move the gripper attached the target item to a post-grasp pose then to the target area along the path computed by the motion planners. 

\section{Perception}
\label{sec:perception}
The perception part of our system takes raw sensor data (point cloud consisting of 307,200 3D points with color) as the input and figures out the full 6 DOF pose of the target object. This procedure consists of three steps: preprocessing, object detection and pose estimation, and the refinement of the estimated pose. These three steps are described in detail below. In addition, we will present the object autoscanning method for automatically generating object templates used in the second step. 

\subsection{Point Cloud Preprocessing}
The raw point cloud is typically dense and noisy, and this will hurt the performance of the perception pipeline. For example, sparse outliers in the point clouds complicate the estimation of local point cloud characteristics and lead to difficulties in computing features of the target object. In our perception pipeline, we first locate the rough bin area containing the target object based on the known shelf pose and use a filter to remove point clouds outside the bin area. Next, the remaining point clouds will be downsampled by a voxelized grid algorithm to reduce the number of points. Then a statistics-based outlier removal algorithm is used to remove noisy measurements from the point clouds. This method use a statistical analysis of each point’s neighborhood to remove points that are far from this point. After preprocessing, we can obtain noise-free, low-density point clouds for object detection and pose estimation. 

\subsection{Object Detection and Pose Estimation}
To detect the target object and estimate the object's 6D pose, we choose LINEMOD~\cite{hinterstoisser2012gradient} as our major method followed by the ICP to refine the pose. As described in~\cite{rennie2015dataset}, LINEMOD is an object detection and pose estimation method, which takes a 3D mesh object model generated in advance as the input. From the model, it combines both 2D (RGB gradients) and 3D (surface normals) information when estimating the full 6D pose of an object. The features are then filtered to a robust set and stored as a template for the object and the given viewpoint. This process is repeated until sufficient coverage of the object is achieved from different viewpoints. These templates are then fed into LINEMOD together with target scene captured by RGBD sensor in real-time. The detection process implements a template matching algorithm followed by several post-processing steps to refine the pose estimate. It uses surface normals in the template matching algorithm and limits RGB gradient features to the object’s silhouette. The result is presented as the coordinate of a specific template in the captured scene, and the template ID indicates the orientation where the coordinate indicates the location. It has open-source implementations in OpenCV~\cite{opencv} and PCL~\cite{pcl}. 

In our tests, we have found that if we use too many templates in the LINEMOD pipeline, it would output undesirable results, e.g., a wrong template in a wrong place. When we reduced the number of template to about 100 templates per object distributed in a hemisphere, the results were much more robust. However, the pose, or more specifically, the orientation of the target item became less accurate in this way, and therefore we need to introduce a registration method for more accurate pose estimation. 

\begin{figure}
\centering
\includegraphics[width=1\linewidth]{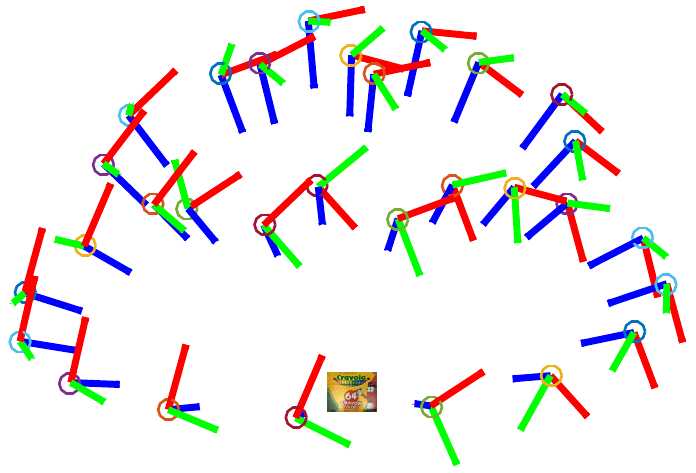}
\caption{The 30 poses for auto-scanning the object.}
\label{fig:autoscan}
\end{figure}

\subsection{Estimated Pose Refinement with ICP} 
In order to improve the accuracy of the estimated pose, we register the resulting template to the point clouds by using the ICP method~\cite{icp}. As shown in Figure~\ref{fig:overview}, the result of pose estimation sometimes may deviate from the point clouds, and the ICP can help to correct such deviations. The refined pose gives us more accurate information about the target object that improves the grasping accuracy and success rate. 

\subsection{Object Autoscanning and Templates Generating}
Since it is tedious, time-consuming and high-priced to manually build the 3D models of the objects, we need to find a way to automatically scan and model the input objects that used in LINEMOD. Moreover, in order to generate enough templates for LINEMOD, we also need an automated mechanism to acquire the templates. The design of the robot holder makes it possible for our robot to scan the object placed under it from different viewpoints. As shown in Figure~\ref{fig:autoscan}, we sample 30 viewpoints on the hemisphere within the working radius of the UR5 robot, that is to say, the robot will move to 30 poses to acquire 3D data of the object. The captured scans can be stitched together with the joint angle information through tracking the transformation of robot coordinate frames over time. 

We can also get the templates directly when we automatically scan the object. Since we use a template matching method for pose estimation, a template of the target item with 6-DOF pose information is what we require. We put the target item on a black acrylic board (which is invisible to RGBD sensor), take snapshots (i.e., point clouds) around it, and use the snapshots directly as templates.

\section{Planning}
\label{sec:planning}
In our system, we compared ITOMP~\cite{park2012itomp}, trajopt~\cite{schulman2014motion}, and other motion planning algorithms in OMPL~\cite{ompl}. None of these algorithms can robustly find a desired path between a pose within a bin and the robot's setup pose. Thus, we divide the planning problem involved in the picking process into two phases: first, we compute a collision-free path for the robot moving from the observation pose to the pre-grasp pose; next, from the pre-grasp pose, we move the gripper to pick up the target object based on a pre-defined grasp plan. Similar to the picking, the placement also has two steps: first, we move the gripper attached with the object from the bin to the post-grasp pose, and then move the gripper to the target location and release the grasped object. As shown in Figure~\ref{fig:overview}, the manipulation pipeline consists of two parts: picking and placement, each part contains motion planning for computing a collision-free trajectory to move the end-effector and grasp planning for computing a valid strategy to place the gripper related to the target object.

\subsection{Motion Planning} 
We integrated ITOMP\footnote{https://youtu.be/xCoaV6sRuTI} into the MoveIt!~\cite{moveit} motion planning framework. For each picking operation, we use this motion planner to compute two collision-free paths, one path is from the observation pose to the pre-grasp pose and the other is from the post-grasp pose to the placement pose. For grasping objects from the shelf, since the pre-grasp and the post-grasp poses are both outside the bin and there is no other obstacle in the workspace, our motion planning framework can successfully compute a collision-free path in most cases. For pick-and-place tasks performed on the table, the post-grasp pose is the same with the pre-grasp pose as there is no collision between the grasped object and the table when moving the gripper away from the bin.

\subsection{Grasp Planning}
For the vacuum gripper, we use a straightforward grasp strategy in both scenarios (the shelf and the table) -- once the perception pipeline obtains the estimated object pose, the robot moves its end-effector to the object until contact, and then picks it up. 

For the M2 gripper and the soft gripper, we generated several predefined grasp using GraspIt~\cite{miller2004graspit}. In GraspIt, we compute the grasp plan for each object based on its 3D model generated by autoscanning as discussed in Section~\ref{sec:perception}. Each pre-computed grasp plan for the object includes an approaching vector and a width between two fingers. After the target item pose is estimated, we first check whether there are any valid approaching vectors based on the collision detection method described in~\cite{pan2012fcl}. If there exists several valid approaching vectors, we choose the one with the highest grasp score as measured by GraspIt, and find a pre-grasp pose where the end-effector is fully outside the bin (or above the table) so that the prismatic joint can stretch the gripper to reach the object. If there is no valid approaching vector, the robot will compute an approaching vector based on the point clouds of the target object and use the gripper to hit the object mildly. Then the robot will repeat the above process until a valid approaching vector is computed.  

Once the pre-grasp pose is determined, the robot moves its end-effector to this pose following the collision-free path computed by motion planners. The robot will extend the gripper along the approaching vector until the target item is within the grasp range. Next, the robot will close the gripper fingers and achieve the grasp of the target object. The gripper with the grasped item will leave the bin or the table along the inverse approaching vector, or first rotate the end-effector to avoid collisions between the object and the bin when moving the gripper away from the bin.

\section{Gripper Design}
\label{sec:gripper}
\begin{figure}
\includegraphics[width=\linewidth]{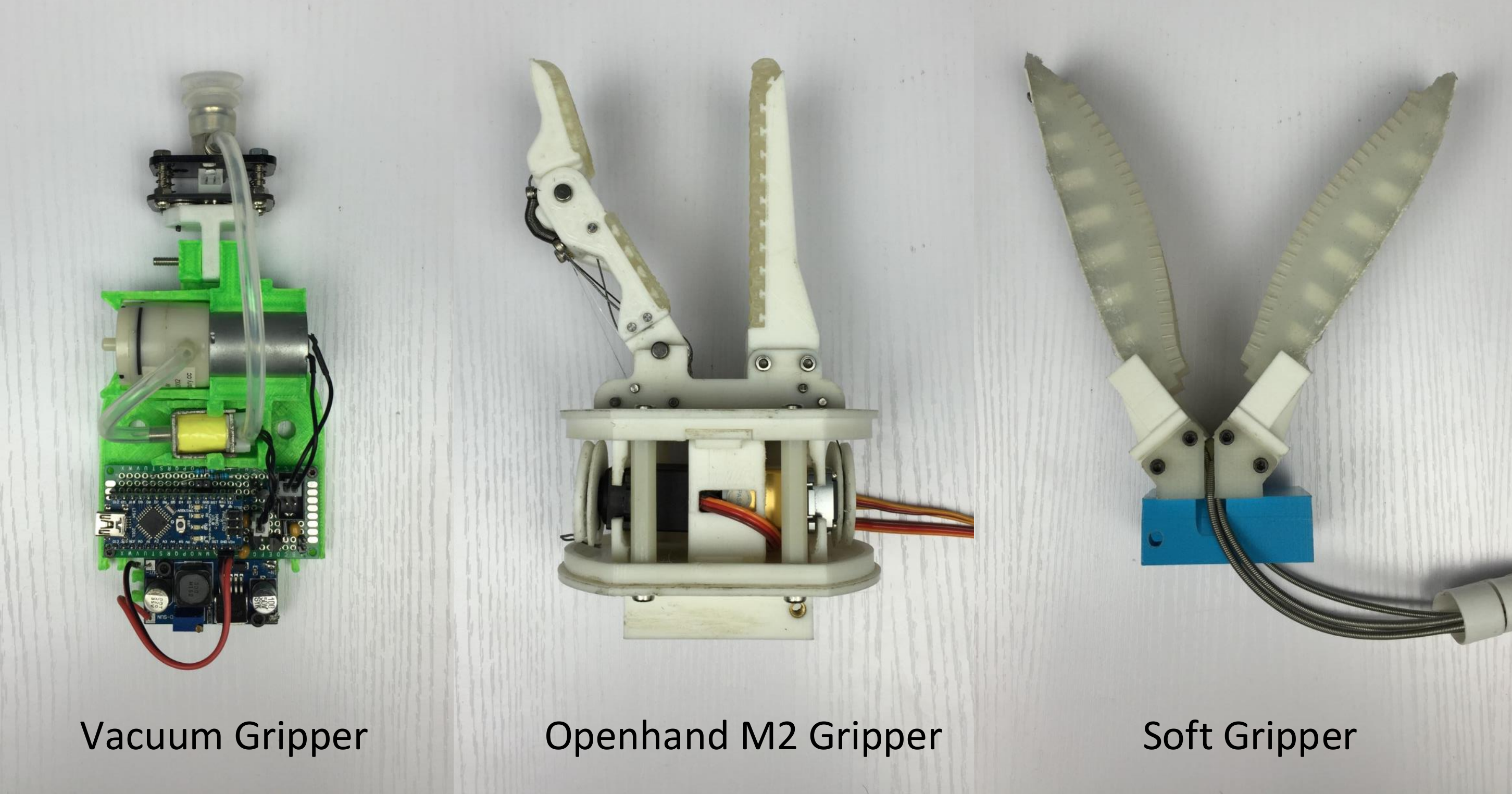}
\caption{Three grippers used in our experiments.}
\label{fig:grippers}
\end{figure}

For the task of picking up general objects with a large variety in terms of shape, material, and weight, we designed and fabricated three different grippers, including a vacuum gripper, the Yale Openhand M2 gripper~\cite{ma2016m2} and the soft gripper, as shown in Figure~\ref{fig:grippers}. None of three grippers can grasp all the 30 objects in Figure~\ref{fig:objects}, but the soft hand can pick up most of objects and the vacuum gripper has good performance on grasping lightweight objects.  

\subsection{Vacuum Gripper}
\label{sec:gripper:vacuum}
We designed a vacuum gripper with pump, valve, electronics and suction cup, mounted on the end-effector of the UR5 robot, powered by the internal power supply of the robot. This vacuum gripper contains a button on the back of the cup to act as a sensor indicating whether the suction cup touched the target object. We tested the vacuum gripper on various objects with varying weight and demonstrated that it can pick up the objects under 300 grams. In the warehouse setting, the vacuum gripper is suitable to pick up items located in the front of the bin or against to the bin sides. Moreover, the gripper's size is too small to get stuck inside a bin, which is an advantage when operating within a small bin and also helps to simplify the grasp planning. Not surprisingly, not all objects under 300 grams can be picked by the suction cup, e.g., the pencil cup holder made of black wire mesh. 

\subsection{Openhand M2 Gripper}
To pick up heavier objects, we built a two-fingered gripper based on the open-source Multi-Modal (M2) Gripper\footnote{It can be accessed online at: \url{http://www.eng.yale.edu/grablab/openhand/model_m2.html}}. This gripper consists of a fixed, modular thumb that can be replaced for different tasks, and a dexterous, tendon-driven index finger that can generate either underactuated or fully-actuated behaviors. This underactuated design extends the grasping capabilities of a simple two-finger gripper since it can produce several distinctive modes of operation. As described in~\cite{ma2016m2}, with only two actuators and basic open-loop control, the hand is able to adaptively grasp objects of varying geometries, pinch-grasp smaller items, and perform some degree of in-hand manipulation via rolling and controlled sliding. 

The M2 gripper can grasp more objects than vacuum cup as shown in Table~\ref{tab:gripper_performance}, including soft or deformable items and the pencil cup. However, it also has several shortcomings: the original design does not have a large enough opening between two fingers, which means it is not able to grasp large objects. Furthermore, if a thin, flat item (such as a thin book) is put against to the bin side, it would be very difficult for the M2 gripper to pick it up even when we equip a ``fingernail" on its tip. It also cannot detect whether it has successfully grasped an item due to lack of a sensor as feedback. Last but not least, the size of the M2 gripper is big for operating inside bins where the robot need to pick the item out of a bin. 

\subsection{Soft Gripper}\label{softgripper}
To overcome the disadvantages of the M2 gripper, we designed a gripper with two soft fingers driven by a cable-sheath system. This gripper has the following features:

\begin{itemize}
\item The distance between two fingertips can be very large when fully opened;
\item It has \emph{fingernails} at its finger tips. When we use it to scoop up the object, the soft property of the finger will adapt it to the uneven surface;
\item The shape of the finger can be roughly determined by the distance of cable pulled, which can be measured by a linear potentiometer;
\item A distance sensor is mounted between two fingers to measure the distance between the target item and the gripper. This sensor can also detect whether there is an item being grasped;
\item The gripper is compact at the fully closed status that allows to work in a relatively small operation space;
\item The gripper is connect to the end-effector of the UR5 robot by a prismatic joint, which allows the gripper to grasp item placed inside a bin.
\end{itemize} 

The experiments presented in the next section demonstrate that this soft gripper can grasp most objects used in our tests. 

\section{Experiments and Discussion}
\label{sec:experiment}
We evaluated our system on various scenarios and used three grippers to grasp assorted items. First, we presented that our system can grasp objects from four corner bins of the shelf. We also evaluated the performance of the robot on grasping the target object from a cluttered bin (e.g., a bin contains four objects). Third, we performed pick-and-place tasks on the table with 30 different objects. Finally, we tested the performance of different grippers for picking up the same objects on the table. 

\subsection{Four Corner Bins}
\begin{figure}
\centering
\includegraphics[width=1\linewidth]{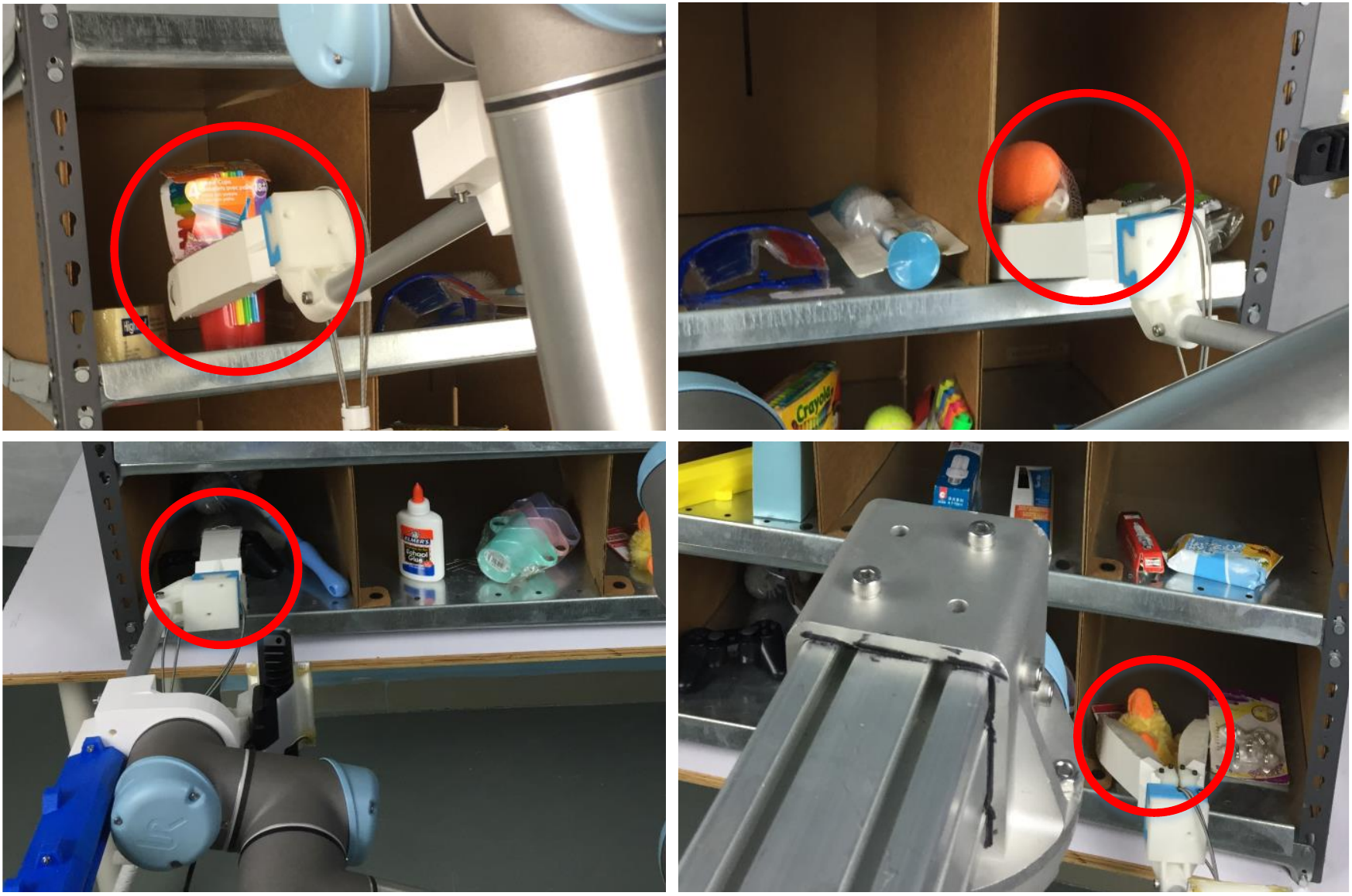}
\caption{Grasping the target object from four corner bins.}
\label{fig:corner_bins}
\end{figure}

Figure~\ref{fig:corner_bins} shows the performance of applying our robot to pick up one target object from four corner bins of the shelf. Based on the motion planning, the robot can follow a collision-free path to access the corner bins. Then relied on the 3D point clouds, the robot estimated 6D pose of the target object, moved its end-effector towards it and grasped the object. In our experiments, the robot can reach all the four corners of the bin. The prismatic joint can even help the soft gripper to pick up objects placed at the deepest location inside the bin.

\subsection{Cluttered Bins}
\begin{figure}
\centering
\includegraphics[width=1\linewidth]{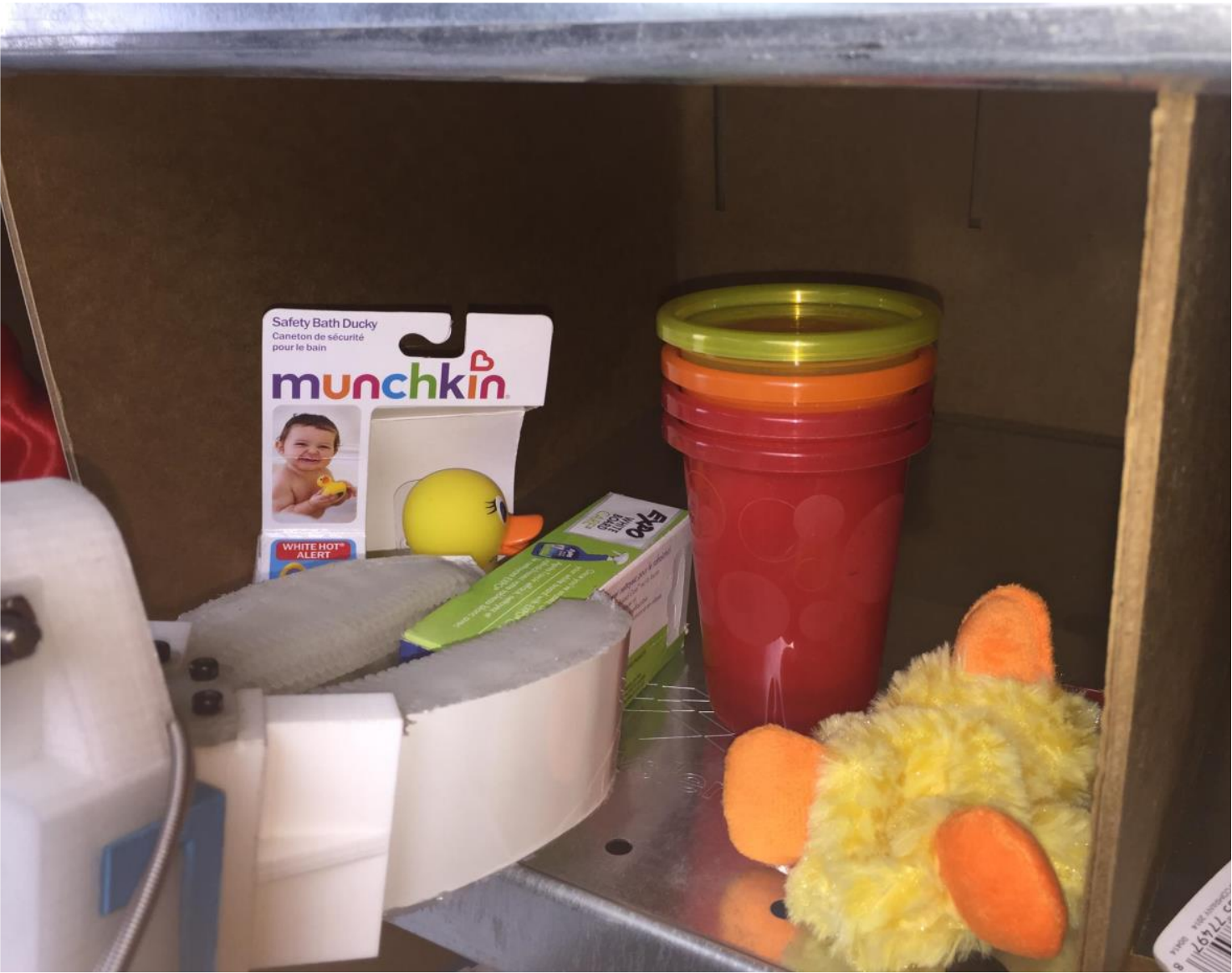}
\caption{Grasping one target object from a cluttered bin.}
\label{fig:cluttered_bin}
\end{figure}

As shown in Figure~\ref{fig:cluttered_bin}, the robot grasped the target object from a cluttered bin containing 4 objects. However, if the target object is occluded by other objects, the RGBD camera will not obtain the 3D data of the object, so the robot cannot pick up the object. Once our perception system find the target object in a cluttered bin, the robot would be able to reliably grasp the object. In addition, it is more reliable for the robot to grasp large objects with distinct color, e.g., Crayola crayons, cat treats, since these objects are more easily detected in a cluttered environment. 

\subsection{Pick-and-place on the Table}
\begin{figure*}
\centering
\includegraphics[width=\linewidth]{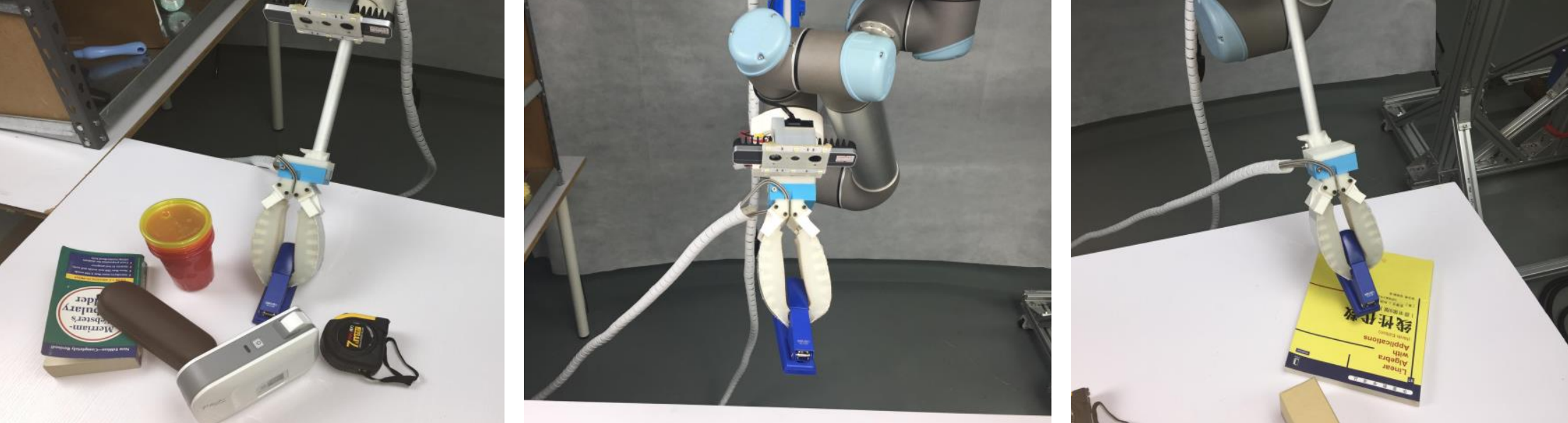}
\caption{Performing pick-and-place on the table.}
\label{fig:pick_n_place}
\end{figure*}

The robot can also perform the common pick-and-place task on the table. We evaluated the performance of the robot on 30 different objects. We divided 30 various objects into 5 groups, each group contained 6 objects disorderly placed on the table. In each evaluation, we provided a list including sorted names of the 6 objects and the specified destinations to our robot, then the robot solved the pick-and-place tasks in sequence. As shown in Figure~\ref{fig:pick_n_place}, the target object was picked up from a clutter on the table and placed at the desired location. 

Again, for our current method, we require the target object to be not occluded by other objects. If most part of the target object is not observed by our RGBD sensor, the perception component will fail to detect and locate the object. Eventually, 26 out of 30 objects were successfully picked by the soft hand. The picture and the sharpie are too thin for the two-fingered gripper to pick up from the table, but they can be easily handled by the vacuum gripper. The cups with transparent bag and the outlet protectors made of transparent material are difficult to detect using our current perception algorithms. 

\subsection{Performance by Different Grippers}
\begin{figure*}
\centering
\includegraphics[width=\linewidth]{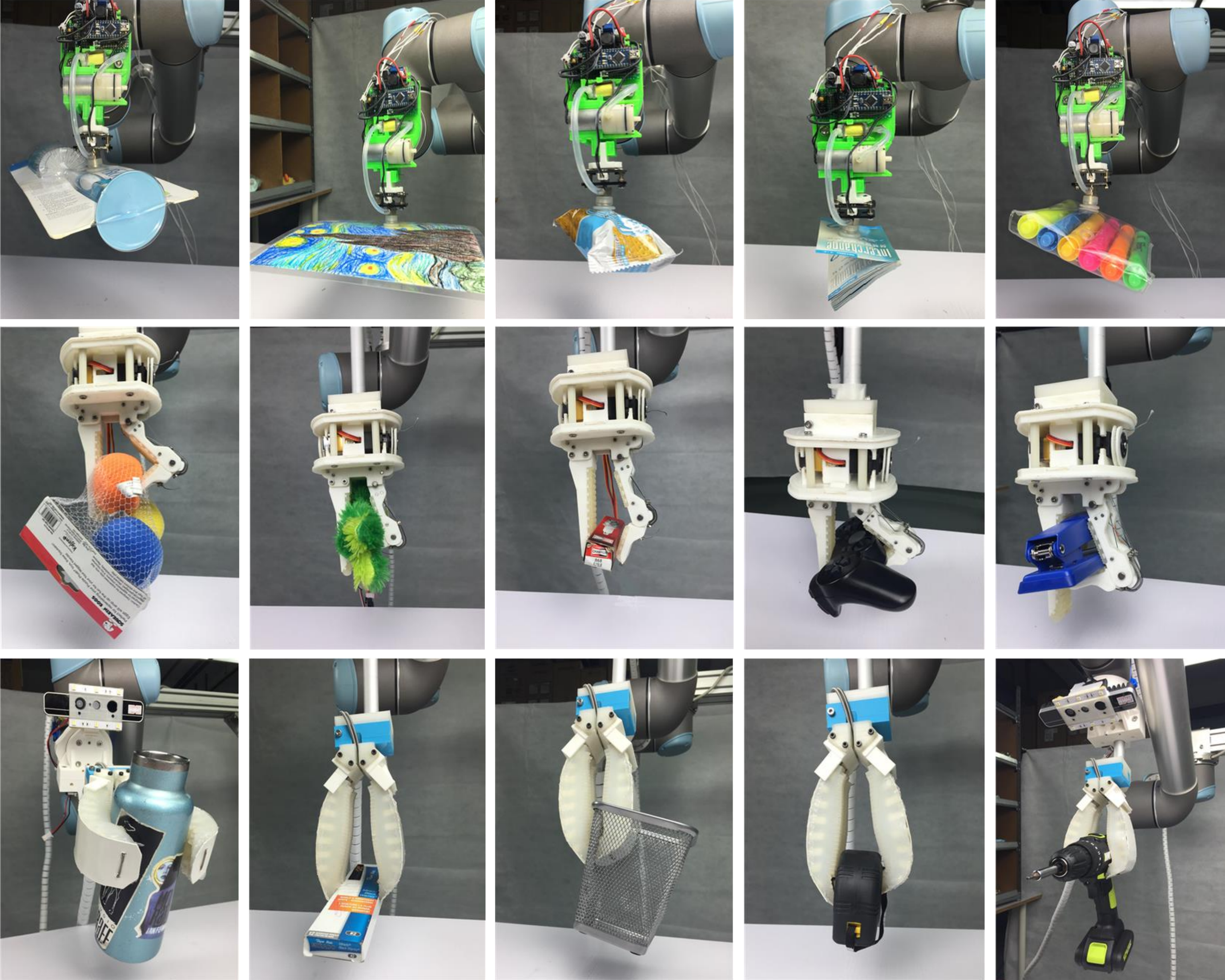}
\caption{Representative objects grasped by our three grippers.}
\label{fig:grasping}
\end{figure*}

Table~\ref{tab:gripper_performance} shows the results of the grasping tests (picking up objects from the table) carried out by the three grippers. It shows that the soft gripper has the best performance on grasping various objects, and the vacuum gripper grasps the fewest objects. This is because, for the vacuum gripper, the successful grasping mostly depends on the object's weight. As we discussed before in Section~\ref{sec:gripper:vacuum}, the vacuum gripper can only grasp objects lighter than 300 grams. If the object is too heavy, the vacuum gripper will fail to pick it up. For the Openhand M2 gripper, it may fail to grasp thin and flat objects as they will twist between the two fingers and cause them to slip out of the two-fingered grasp during the movement. However, this situation rarely happens to the soft gripper due to the high-damping, high-friction property of its fingers. Figure~\ref{fig:grasping} shows the representative objects grasped by these three grippers.

\begin{table}
\centering
\begin{tabular}{l|c|c|c}
Object & Vacuum Gripper & Openhand M2 & Soft Gripper \\ 
\hline
\hline
Cardboard box & ${\surd}$ & ${\times}$ & ${\surd}$  \\
Spark plug & ${\surd}$ & ${\surd}$ & ${\surd}$ \\
Crayola crayons & ${\surd}$ & ${\surd}$ & ${\surd}$ \\
Whiteboard eraser & ${\surd}$ & ${\surd}$ & ${\surd}$ \\
Glue & ${\surd}$ & ${\surd}$ & ${\surd}$ \\
Sharpie & ${\surd}$ & ${\times}$ & ${\times}$ \\
Cat treats & ${\surd}$ & ${\surd}$ & ${\surd}$ \\
Pencils & ${\surd}$ & ${\surd}$ & ${\surd}$ \\
Electric drill & ${\times}$ & ${\surd}$ & ${\surd}$ \\
Screwdrivers & ${\surd}$ & ${\times}$ & ${\surd}$ \\
Foam balls & ${\times}$ & ${\surd}$ & ${\surd}$ \\
Bottle cleaner & ${\surd}$ & ${\surd}$ & ${\surd}$ \\
Dog toy - frog & ${\surd}$ & ${\surd}$ & ${\surd}$ \\
Dog toy - ball & ${\surd}$ & ${\surd}$ & ${\surd}$ \\
Outlet protectors & ${\times}$ & ${\times}$ & ${\times}$ \\
Cookie & ${\surd}$ & ${\surd}$ & ${\surd}$ \\
Picture & ${\surd}$ & ${\times}$ & ${\times}$ \\
Bottle & ${\times}$ & ${\surd}$ & ${\surd}$ \\
Dictionary & ${\surd}$ & ${\surd}$ & ${\surd}$ \\
Plastic cups & ${\surd}$ & ${\surd}$ & ${\surd}$ \\
Gamepad & ${\surd}$ & ${\surd}$ & ${\surd}$ \\
Little book & ${\surd}$ & ${\surd}$ & ${\surd}$ \\
Pocket ruler & ${\times}$ & ${\surd}$ & ${\surd}$ \\
Sticky notes & ${\surd}$ & ${\surd}$ & ${\surd}$ \\
Pencil holder & ${\times}$ & ${\surd}$ & ${\surd}$ \\
3D printed model & ${\surd}$ & ${\surd}$ & ${\surd}$ \\
Cups with package & ${\times}$ & ${\times}$ & ${\times}$ \\
Earplug box & ${\surd}$ & ${\surd}$ & ${\surd}$ \\
Rubber ducky & ${\surd}$ & ${\surd}$ & ${\surd}$ \\
\end{tabular}
\caption{Performance of three grippers on grasping 30 objects from the table.}
\label{tab:gripper_performance} 
\end{table}

\subsection{Failure Cases}
\begin{figure}
\centering
\includegraphics[width=\linewidth]{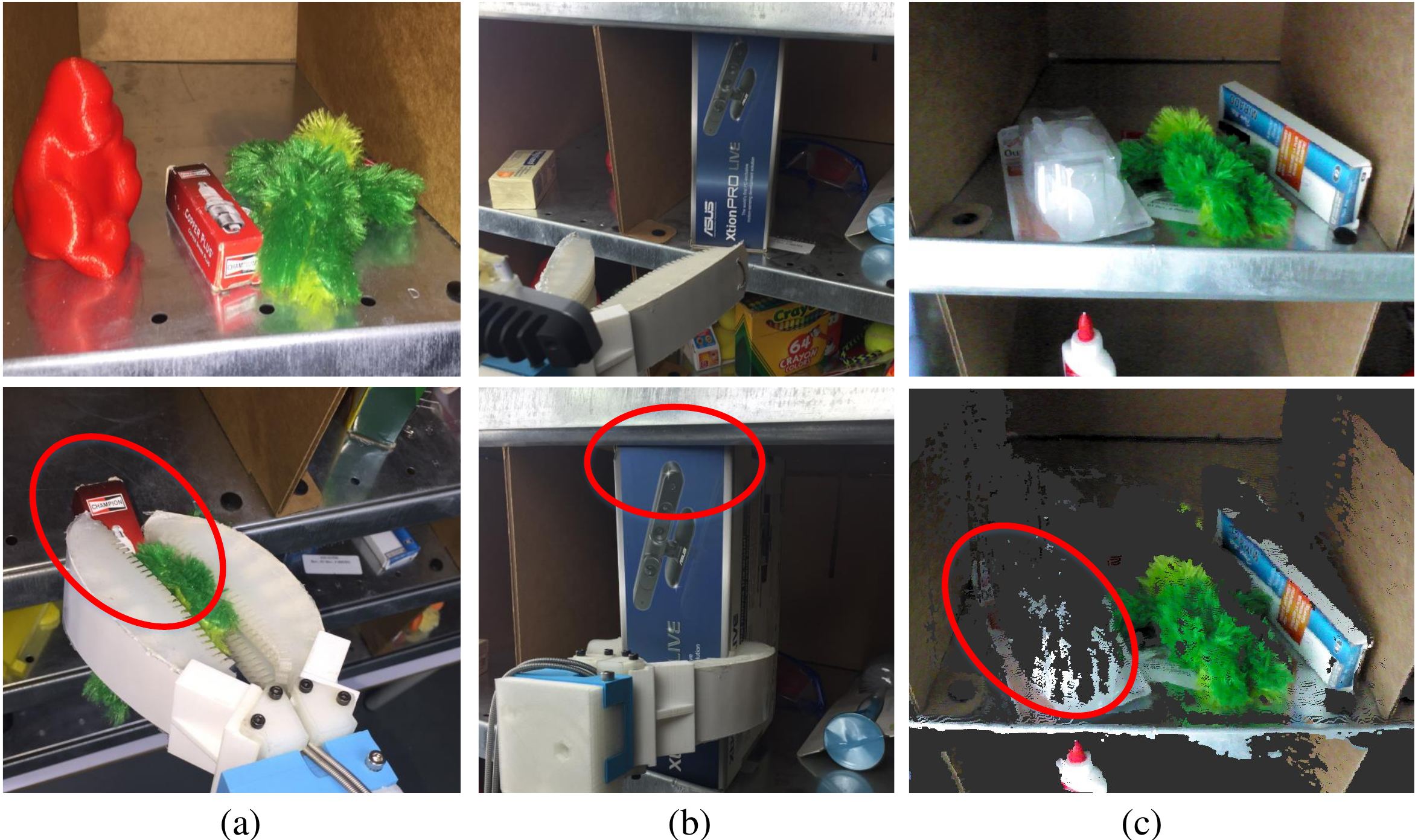}
\caption{Typical failure cases: (a) simultaneously picking up a target item and a non-target item; (b) getting stuck in the bin; (c) no point clouds for the transparent object}
\label{fig:failure_cases}
\end{figure}
In the evaluation we met several interesting failure cases. As shown in Figure~\ref{fig:failure_cases}(a), if the target object is very close to other objects, the robot may simultaneously pick up the target object with a non-target item. This is because our current pipeline does not have a scheme to separate them. In addition, while picking up a very large object from the bin as shown in Figure~\ref{fig:failure_cases}(b), the robot may get stuck inside the bin. Furthermore, the perception method proposed in this paper currently has a low precision for transparent or reflective objects as they are invisible to our RGBD camera as shown in Figure~\ref{fig:failure_cases}(c). As a result, object material properties play an important role in the success rate of our perception algorithm. 

\subsection{Lessons Learned}
\label{sec:lesson}
To build an autonomous robot that can pick and place general objects is a systematic project involving hardware, software and algorithmic components in robotics. It also requires a combination of state-of-the-art in robotics research community with expertise already existing in industry, in order to build a task-specific, reliable and integrated system. The following are lessons we learned during building this picking system: (i) In the warehouse setting, a well-designed, task-specific perception method outperforms the general-purpose approaches, and simple, heuristic approaches may be a more promising route to design perception systems for specific tasks. Thus, the key challenge is how to incorporate information about the specific scenario into generic state-of-the-art approaches developed in the research community. (ii) Hardware and software are a unity, two sides of one coin. When we build an integrated system, we face both sides simultaneously. For example, different grippers require different grasp planning algorithms, i.e., the hardware of grippers, to some extension, decides the complexity of grasp planning methods. Thus, it is critical to making a task-oriented trade-off between the hardware and software. Moreover, the hardware design should follow the structure characteristic of the workspace, e.g., a smaller gripper has obvious advantages over the ``big" gripper for manipulation in tight spaces. (iii) An error recovery mechanism is necessary for building a reliable system. Dealing with the errors such as dropped objects, destroyed objects, or miss-classified objects plays a key role in implementing warehouse automation.

\section{Conclusion}
\label{sec:conclusion}
We presented and evaluated our system to pick and place general objects in a simplified warehouse setting. This work demonstrated how to build an autonomous robot that can pick general objects from the shelves and perform the pick-and-place tasks on the table. We believe that the system presented in this paper is a necessary step to build an automated warehouse system. For the future work, we plan to update our system into a mobile manipulator which can autonomously move inside the warehouse, to refine the mechanical design of the soft gripper to deal with objects with more and more variety, and to improve the performance of our perception algorithms on the transparent and reflective objects.

\bibliographystyle{IEEEtran}
\bibliography{references}
\end{document}